# A FUZZY AHP APPROACH FOR SUPPLIER SELECTION PROBLEM: A CASE STUDY IN A GEARMOTOR COMPANY

## Mustafa Batuhan AYHAN[1]

### Department of Industrial Engineering, Marmara University, Istanbul, TURKEY

## ABSTRACT:

*Supplier selection is one of the most important functions of a purchasing department. Since by deciding the best supplier, companies can save material costs and increase competitive advantage. However this decision becomes complicated in case of multiple suppliers, multiple conflicting criteria, and imprecise parameters. In addition the uncertainty and vagueness of the experts' opinion is the prominent characteristic of the problem. Therefore an extensively used multi criteria decision making tool Fuzzy AHP can be utilized as an approach for supplier selection problem. This paper reveals the application of Fuzzy AHP in a gear motor company determining the best supplier with respect to selected criteria. The contribution of this study is not only the application of the Fuzzy AHP methodology for supplier selection problem, but also releasing a comprehensive literature review of multi criteria decision making problems. In addition by stating the steps of Fuzzy AHP clearly and numerically, this study can be a guide of the methodology to be implemented to other multiple criteria decision making problems.*

## Keywords:

## 1. INTRODUCTION

Supplier selection, which includes multi criteria and multiple conflicting objectives, can be defined as the process of finding the right suppliers with the right quality at the right price, at the right time, and in the right quantities. It is noted that, manufacturers spend more than 60% of its total sales on purchased items [1]. In addition, their purchases of goods and services constitute up to 70% of product cost [2]. Therefore, selecting the right supplier significantly reduces purchasing costs, improves competitiveness in the market and enhances end user satisfaction [3]. Since this selection process mainly involves the evaluation of different criteria and various supplier attributes, it can be considered as a multiple criteria decision making (MCDM) problem [4]. Based on several criteria and alternatives to be considered, various decision making methods have been proposed to provide a solution to this problem [5].

Basically there are two types of supplier selection problems [6]. In single sourcing type, one supplier can satisfy all the buyer's needs. In the multiple sourcing type, no supplier can satisfy all the buyer's requirements. Hence the management wants to split order quantities among different suppliers [7].

As a pioneer in the supplier selection problem, Dickson [8] identified 23 different criteria for selecting suppliers, including quality, delivery, performance history, warranties, price, technical capability, and financial position [9]. With a thorough literature survey, Weber, et al. [10] reviewed 74 different articles by classifying into three categories; linear weighting methods, mathematical programming models, and statistical approaches. Following Weber et al. [10], De Boer et al. [11], identified four stages for supplier selection including; definition of the problem, formulation of criteria, qualification, and final selection respectively [12].

According to one of the recent classifications made by Sanayei et al. [13], there are six classes. These are multi attribute decision making techniques (Analytic Hierarchy Process- AHP, Analytic Network Process- ANP, Technique for Order Preference by Similarity to Ideal Solution- TOPSIS), mathematical programming (Linear Programming- LP, Goal Programming- GP or Mixed Integer Programming- MIP), probabilistic approaches, intelligent approaches (neural networks, expert systems), hybrid approaches (AHP-LP, ANP-MIP) and others.

This study is mapped as; the literature is reviewed according to the different criteria and methods used for the supplier selection problem in the second part. Part 3 explains the Fuzzy AHP method in detail which is utilized to solve the supplier selection problem of a manufacturing firm elaborated as a case study in the fourth part. Part 5 presents the conclusion and directs for further steps of this study with the references following.

## 2. LITERATURE  REVIEW

As mentioned previously there are comprehensive literature reviews performed before such as Dickson [8], Weber et al. [10], De Boer et al. [11] and Sanayei et al. [13]. However, in this part, at first, the literature will be reviewed according to the selection criteria and then the methodologies used for supplier selection problem will be explained mainly based on a previous study performed by Ayhan [14].

Many studies have been performed by using different criteria starting from the Dickson's 23 criteria [8]. Cheraghi et al. [15] updated Dickson's criteria with 13 more and stated that as the pace of market globalization quickens, the number of criteria to be considered will increase [16]. As a brief of all criteria that have appeared in literature since 1966, quality, price, and delivery performances are suggested as the most important selection criteria [4].

When the methodologies used for solving supplier selection problem are reviewed, it is observed that, various multi criteria decision making methods are implemented, which can be grouped into three broad categories [17].

1) *Value Measurement Models:* AHP and multi attribute utility theory (MAUT) are the best known method in this group.
2) *Goal, Aspiration, and Reference Models:* Goal programming and TOPSIS are the most important methods that belong to the group.
3) *Outranking Methods:* ELECTRE and PROMETHEE are two main families of methods in this group.

AHP, which was first developed by Saaty [18], integrates experts' opinions and evaluation scores into a simple elementary hierarchy system by decomposing complicated problems from higher hierarchies to lower ones. Yahya and Kingsman [19] are one of the first known researchers to use AHP to determine priorities in selecting suppliers. Similarly Analytic Network Process (ANP) is also a multi attribute approach for decision making that allows the transformation of qualitative values to quantitative ones. Since AHP is a special case of ANP and it does not contain feedback loops among the factors, ANP is used to determine supplier selection for the longer terms [3].

However since the uncertainty and vagueness of the experts' opinion is the prominent characteristic of the problem, this impreciseness of human's judgments can be handled through the fuzzy sets theory developed by Zadeh [20]. Fuzzy AHP method [21], [22], [23] systematically solves the selection problem that uses the concepts of fuzzy set theory and hierarchical structure analysis. Basically, Fuzzy AHP method represents the elaboration of a standard AHP method into fuzzy domain by using fuzzy numbers for calculating instead of real numbers [24]. On the other hand, since ANP deals only crisp comparison ratios, uncertain human judgments can be dealt with Fuzzy ANP, in which the weights are simpler to calculate than for conventional ANP [3].

In case of many pair wise comparisons, ANP, AHP, FAHP, or FANP becomes burdensome to cope with. Instead TOPSIS, which is a widely accepted multi attribute decision making tool can be used [25]. The concept of TOPSIS is that the most preferred alternative should not only have the shortest distance from the positive ideal solution, but should also be farthest from the negative ideal solution [17]. Chen et al. [26] extended the concept of TOPSIS to fuzzy environments by using fuzzy linguistic values. This fuzzy TOPSIS method fits human thinking under actual environment.

Furthermore ELECTRE (Elimination et Choice Translating Reality), which was first introduced by Benayoun et al. [27], concerns the concordance, discordance and out ranking concepts originating from real world applications. ELECTRE methods have been applied to problems in many areas including energy [28], environment management [29], finance [30], project selection [31], and decision analysis [32]. Details and the derivatives of ELECTRE method can be found in the literature [33].

In addition, the PROMETHEE method (Preference Ranking Organization Method for Enrichment Evaluations) is one of the most recent MCDA methods that was developed by Brans [34] and further extended by Vincke and Brans [35]. PROMETHEE is an outranking method for a finite set of alternative actions to be ranked and selected among criteria, which are often conflicting. PROMETHEE is also a quite simple ranking method in conception and application compared with the other methods for multi-criteria analysis [36]. Since the main focus of this paper is only limited to application of Fuzzy AHP, a comprehensive literature review on methodologies and applications of PROMETHEE can be found in the literature [37].

Although there are many applications of F-AHP in various fields including; personnel selection [38], weapon selection [39], energy alternatives selection [40], job selection [41] and performance evaluation systems [42], [43] only the recent Fuzzy AHP applications for supplier selection problems will be elaborated in forthcoming paragraphs.

In 2010, a Fuzzy AHP method is used for supplier selection in electronic market places [44]. According to their two phase methodology, at the first phase, initial screening of the suppliers through the enforcement of hard constraints on the selection criteria is performed. In the second phase, final supplier evaluation is performed through the application of a modified variant of Fuzzy AHP. This methodology facilitates an easier elicitation of user preferences through the reduction of necessary user input (i.e. pair wise comparisons) and reduces computational complexity.

In 2011, Fuzzy AHP approach is used for supplier selection in a washing machine company [45]. First they determine the criteria providing the most customer satisfaction and design the hierarchy structure including the main attributes and sub-attributes for supplier selection. The weights of the attributes and alternatives are calculated using pair wise comparison matrices.

In 2012, a combination of fuzzy AHP and fuzzy objective linear programming is used to select the best supplier to develop a low carbon supply chain [46]. At first, Fuzzy AHP is used to determine weights of predetermined criteria, which are cost, quality, rejection percentage, late delivery percentage, green house gas emission and demand. Then, by the help of fuzzy objective linear programming, the best supplier is determined.

In 2013, an interactive solution approach is proposed for multiple objective supplier selection problems with Fuzzy AHP [16]. Their methodology includes three objectives; minimizing total monetary cost, maximizing total quality and maximizing service level. By the provided interactivity, the decision maker has the opportunity to incorporate his preferences during the iterations of the optimization process.

Based on comprehensive literature review, considering multi criteria structure of the supplier selection problem and the vagueness in real environment, fuzzy AHP is thought to be a suitable and simple enough for selecting the best supplier. In the next section the details of Fuzzy AHP is given in detail.

## 3. FUZZY ANALYTIC HIERARCHY PROCESS (F-AHP)

Fuzzy Analytic Hierarchy Process (F-AHP) embeds the fuzzy theory to basic Analytic Hierarchy Process (AHP), which was developed by Saaty [18]. AHP is a widely used decision making tool in various multi-criteria decision making problems. It takes the pair-wise comparisons of different alternatives with respect to various criteria and provides a decision support tool for multi criteria decision problems. In a general AHP model, the objective is in the first level, the criteria and sub criteria are in the second and third levels respectively. Finally the alternatives are found in the fourth level [45].

Since basic AHP does not include vagueness for personal judgments, it has been improved by benefiting from fuzzy logic approach. In F-AHP, the pair wise comparisons of both criteria and the alternatives are performed through the linguistic variables, which are represented by triangular numbers [45]. One of the first fuzzy AHP applications was performed by van Laarhoven and Pedrycz [47]. They defined the triangular membership functions for the pair wise comparisons. Afterwards, Buckley [48] has contributed to the subject by determining the fuzzy priorities of comparison ratios having triangular membership functions. Chang [49] also introduced a new method related with the usage of triangular numbers in pair-wise comparisons. Although there are some more techniques embedded in F-AHP, within the scope of this study, Buckley's methods [48] is implemented to determine the relative importance weights for both the criteria and the alternatives. The steps of the procedure are as follows:

**Step 1:** Decision Maker compares the criteria or alternatives via linguistic terms shown in Table 1.

Table 1: Linguistic terms and the corresponding triangular fuzzy numbers

| Saaty scale | Definition | Fuzzy Triangular Scale |
|---|---|---|
| 1 | Equally important (Eq. Imp.) | (1, 1, 1) |
| 3 | Weakly important (W. Imp.) | (2, 3, 4) |
| 5 | Fairly important (F. Imp.) | (4, 5, 6) |
| 7 | Strongly important (S. Imp.) | (6, 7, 8) |
| 9 | Absolutely important (A. Imp.) | (9, 9, 9) |
| 2 | | (1, 2, 3) |
| 4 | The intermittent values between two adjacent scales | (3, 4, 5) |
| 6 | | (5, 6, 7) |
| 8 | | (7, 8, 9) |

According to the corresponding triangular fuzzy numbers of these linguistic terms, for example if the decision maker states "Criterion 1 (C1) is Weakly Important than Criterion 2 (C2)", then it takes the fuzzy triangular scale as (2, 3, 4). On the contrary, in the pair wise contribution matrice of the criteria, comparison of C2 to C1 will take the fuzzy triangular scale as (1/4, 1/3, 1/2).

The pair wise contribution matrice is shown in Eq.1, where $\widetilde{d_{ij}^{k}}$ indicates the k[th] decision maker's preference of i[th] criterion over j[th] criterion, via fuzzy triangular numbers. Here, "tilde" represents the triangular number demonstration and for the example case, $\widetilde{d_{12}^{1}}$ represents the first decision maker's preference of first criterion over second criterion, and equals to, $\widetilde{d_{12}^{1}} = (2, 3, 4)$.

$$\widetilde{A}^{k} = \begin{bmatrix} \widetilde{d}_{11}^{k} & \widetilde{d}_{12}^{k} & \dots & \widetilde{d}_{1n}^{k} \\ \widetilde{d}_{21}^{k} & \dots & \dots & \widetilde{d}_{2n}^{k} \\ \dots & \dots & \dots & \dots \\ \widetilde{d}_{n1}^{k} & \widetilde{d}_{n2}^{k} & \dots & \widetilde{d}_{nn}^{k} \end{bmatrix} \tag{1}$$

**Step 2:** If there is more than one decision maker, preferences of each decision maker $(\widetilde{d_{ij}^{k}})$ are averaged and $(\widetilde{d_{ij}})$ is calculated as in the Eq. 2.

$$\widetilde{d}_{ij} = \frac{\sum_{k=1}^{K} \widetilde{d}_{ij}^{k}}{K} \tag{2}$$

**Step 3:** According to averaged preferences, pair wise contribution matrice is updated as shown in Eq. 3.

$$\widetilde{A} = \begin{bmatrix} \widetilde{d_{11}} & \cdots & \widetilde{d_{1n}} \\ \vdots & \ddots & \vdots \\ \widetilde{d_{n1}} & \cdots & \widetilde{d_{nn}} \end{bmatrix} \tag{3}$$

**Step 4:** According to Buckley [48], the geometric mean of fuzzy comparison values of each criterion is calculated as shown in Eq. 4. Here, $\widetilde{r}_{i}$ still represents triangular values.

$$\widetilde{r}_{i} = \left( \prod_{j=1}^{n} \widetilde{d}_{ij} \right)^{1/n}, \quad i=1, 2,...,n \tag{4}$$

**Step 5:** The fuzzy weights of each criterion can be found with Eq. 5, by incorporating next 3 sub steps.

**Step 5a**: Find the vector summation of each $\widetilde{r}_{i}$.

**Step 5b**: Find the (-1) power of summation vector. Replace the fuzzy triangular number, to make it in an increasing order.

**Step 5c**: To find the fuzzy weight of criterion i $(\widetilde{w}_{i})$, multiply each $\widetilde{r}_{i}$ with this reverse vector.

$$\begin{aligned} \widetilde{w}_{i} &= \widetilde{r}_{i} \otimes \left( \widetilde{r}_{1} \oplus \widetilde{r}_{2} \oplus \cdots \oplus \widetilde{r}_{n} \right)^{-1} \\ &= \left( lw_{i}, mw_{i}, uw_{i} \right) \end{aligned} \tag{5}$$

**Step 6:** Since $\widetilde{w_t}$ are still fuzzy triangular numbers, they need to de-fuzzified by Centre of area method proposed by Chou and Chang [50], via applying the equation 6.

$$M_i = \frac{lw_i + mw_i + uw_i}{3} \tag{6}$$

**Step 7:** $M_i$ is a non fuzzy number. But it needs to be normalized by following Eq. 7.

$$N_i = \frac{M_i}{\sum_{i=1}^{n} M_i} \tag{7}$$

These 7 steps are performed to find the normalized weights of both criteria and the alternatives. Then by multiplying each alternative weight with related criteria, the scores for each alternative is calculated. According to these results, the alternative with the highest score is suggested to the decision maker. In order to make the methodology clear and see its applicability, a real case study in a gear motor company is revealed in the next section.

## 4. APPLICATION IN A GEARMOTOR COMPANY

The Fuzzy AHP methodology is applied in a gear motor company which produces frequency inverters and decentralized Drive Engineering motors in Turkey. In fact, previously a study has been performed to find the best supplier for this company by Fuzzy TOPSIS method, the technical details of the firm can be found in the literature [14]. In order to keep the business confidentiality, the name of the company and the alternative suppliers are preserved. Base on the previous study, "bear ring", which is the most frequently used raw material, taken into account to determine the best supplier among 3 alternative suppliers and regarding 5 criteria. The main frame of the supplier selection for the related company can be represented as following Figure 1. Here, both the criteria and the alternative weights should be calculated. Therefore, these two parts will be analyzed separately.

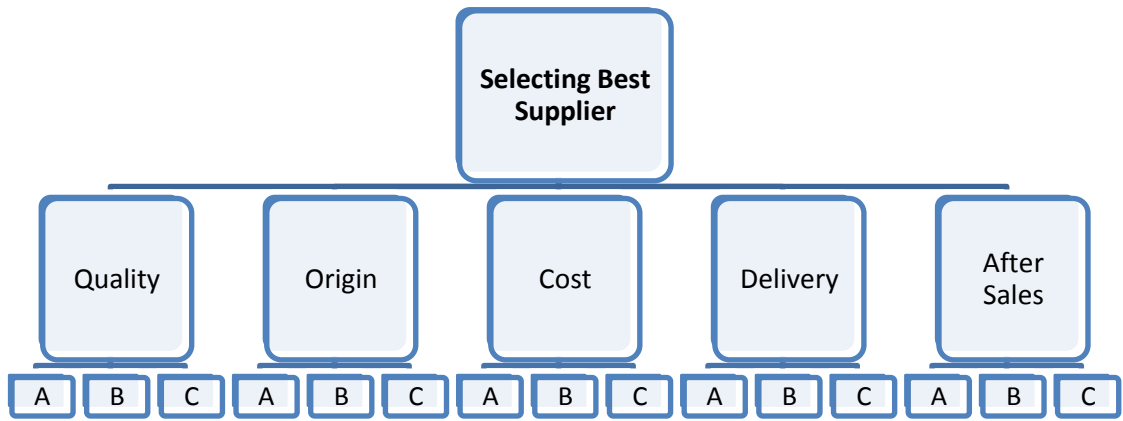

Figure 1: The hierarchy of the criteria and the alternatives

## 4.1    Determining Weights of Criteria

In order to determine the criteria and evaluate the alternatives for the supplier selection process, a meeting was performed with both the production manager and purchasing manager. According to their preferences, the averaged pair wise comparison of the criteria is represented by following Table 2.

Table 2: Pair Wise Comparisons of Criteria

| Q # | A. Imp. (9, 9, 9) | S. Imp. (6, 7, 8) | F. Imp. (4, 5, 6) | W. Imp. (2, 3, 4) | CRITERION | Eq. Imp. (1, 1, 1) | CRITERION | W. Imp. (2, 3, 4) | F. Imp. (4, 5, 6) | S. Imp. (6, 7, 8) | A. Imp. (9, 9, 9) |
|---|---|---|---|---|---|---|---|---|---|---|---|
| 1 | | | | | QUALITY | ✓ | ORIGIN | | | | |
| 2 | | | ✓ | | QUALITY | | COST | | | | |
| 3 | | ✓ | | | QUALITY | | DELIVERY | | | | |
| 4 | | | ✓ | | QUALITY | | AFTER SALES | | | | |
| 5 | | | ✓ | | ORIGIN | | COST | | | | |
| 6 | | ✓ | | | ORIGIN | | DELIVERY | | | | |
| 7 | | ✓ | | | ORIGIN | | AFTER SALES | | | | |
| 8 | | | | | COST | | DELIVERY | ✓ | | | |
| 9 | | | | ✓ | COST | | AFTER SALES | | | | |
| 10 | | | | | DELIVERY | | AFTER SALES | | ✓ | | |

According to Table 2, pair wise comparison matrice is formed as Table 3;

Table 3: Comparison matrice for criteria

| CRITERIA | Quality | Origin | Cost | Delivery | After Sales |
|---|---|---|---|---|---|
| **Quality** | (1,1,1) | (1,1,1) | (4,5,6) | (6,7,8) | (4,5,6) |
| **Origin** | (1,1,1) | (1,1,1) | (4,5,6) | (6,7,8) | (6,7,8) |
| **Cost** | (1/6,1/5,1/4) | (1/6,1/5,1/4) | (1,1,1) | (1/4,1/3,1/2) | (2,3,4) |
| **Delivery** | (1/8,1/7,1/6) | (1/8,1/7,1/6) | (2,3,4) | (1,1,1) | (1/6,1/5,1/4) |
| **After Sales** | (1/6,1/5,1/4) | (1/8,1/7,1/6) | (1/4,1/3,1/2) | (4,5,6) | (1,1,1) |

After completing the first three steps of the methodology, at the fourth step, the geometric mean of fuzzy comparison values of each criterion is calculated by Eq. 4. For example, $\tilde{r}_1$-geometric mean of fuzzy comparison values of 'Quality' criterion is calculated as Eq. 8;

$$\tilde{r}_i = \left(\prod_{j=1}^{n} \tilde{d}_{ij}\right)^{1/n} = \left[(1*1*4*6*)^{\frac{1}{5}} ; (1*1*5*7*5)^{\frac{1}{5}}; \left(1*1*6*8*6^{\frac{1}{5}}\right)\right]$$
$$= [2.49 ; 2.81; 3.10] \tag{8}$$

Hence, the geometric means of fuzzy comparison values of all criteria are shown in Table 4. In addition, the total values and the reverse values are also presented. In the last row of Table 4, since the fuzzy triangular number should be in increasing order, the order of the numbers is changed.

Table 4: Geometric means of fuzzy comparison values

| CRITERIA | $\tilde{r}_i$ | | |
|---|---|---|---|
| Quality | 2.49 | 2.81 | 3.10 |
| Origin | 2.70 | 3.00 | 3.29 |
| Cost | 0.43 | 0.53 | 0.66 |
| Delivery | 0.35 | 0.41 | 0.49 |
| After Sales | 0.46 | 0.54 | 0.66 |
| Total | 6.43 | 7.30 | 8.20 |
| Reverse (power of -1) | 0.16 | 0.14 | 0.12 |
| Increasing Order | 0.12 | 0.14 | 0.16 |

In the fifth step, the fuzzy weight of 'Quality' criterion ($\widetilde{w_1}$) is found by the help of Eq. 5 and shown in Eq. 9

$$\widetilde{w_1} = [(2.49 * 0.12); (2.81 * 0.14); (3.10 * 0.16)] = [0.304; 0.385; 0.483] \tag{9}$$

Hence the relative fuzzy weights of each criterion are given in Table 5;

Table 5: Relative fuzzy weights of each criterion

| CRITERIA | $\widetilde{w_i}$ | | |
|---|---|---|---|
| Quality | 0.304 | 0.385 | 0.483 |
| Origin | 0.330 | 0.412 | 0.511 |
| Cost | 0.052 | 0.072 | 0.103 |
| Delivery | 0.043 | 0.057 | 0.076 |
| After Sales | 0.056 | 0.075 | 0.103 |

In the sixth step, the relative non-fuzzy weight of each criterion ($M_i$) is calculated by taking the average of fuzzy numbers for each criterion. In the seventh step, by using non fuzzy $M_i$'s, the normalized weights of each criterion are calculated and tabulated in Table 6.

Table 6: Averaged and normalized relative weights of criteria

| CRITERIA | $M_i$ | $N_i$ |
|---|---|---|
| Quality | 0.391 | 0.383 |
| Origin | 0.418 | 0.409 |
| Cost | 0.075 | 0.074 |
| Delivery | 0.058 | 0.057 |
| After Sales | 0.078 | 0.076 |

## 4.2    Determining Weights of Alternatives with respect to Criteria

After achieving the normalized non-fuzzy relative weights for criteria, the same methodology is applied to find the respective values for alternatives. But now, the alternatives should be pair wise compared with respect to each criterion particularly. That means, this analysis should be repeated for 5 more times for each criterion. However, it will be burdensome to explain for each 5 of them; only "Quality" criterion will be handled.

Pair wise comparison of alternatives with respect to "Quality" criterion is interviewed and the following Table 7 is achieved.

Table 7: Pair Wise Comparisons of Alternatives with respect to "Quality" Criteria

| Q # | A. Imp. (9,9,9) | S. Imp. (6,7,8) | F. Imp. (4,5,6) | W. Imp. (2,3,4) | ALTERNATİVES | Eq. Imp. (1,1,1) | ALTERNATİVES | W. Imp. (2,3,4) | F. Imp. (4,5,6) | S. Imp. (6,7,8) | A. Imp. (9,9,9) |
|---|---|---|---|---|---|---|---|---|---|---|---|
| 1 | | | | | A1 | | A2 | | ✓ | | |
| 2 | | | | | A1 | | A3 | | | | ✓ |
| 3 | | | | | A2 | | A3 | ✓ | | | |

According to Table 7; pair wise comparison matrice is formed as Table 8;

Table 8: Comparison matrice of alternatives with respect to "Quality" criterion

| ALTERNATIVES | A1 | A2 | A3 |
|---|---|---|---|
| A1 | (1,1,1) | (1/6,1/5,1/4) | (1/9,1/9,1/9) |
| A2 | (4,5,6) | (1,1,1) | (1/4,1/3,1/2) |
| A3 | (9,9,9) | (2,3,4) | (1,1,1) |

Similar to criterion calculation methodology, the geometric means of fuzzy comparison values ($\widetilde{r_i}$) and relative fuzzy weights of alternatives for each criterion ($\widetilde{w_i}$) are tabulated in Table 9.

Table 9: Geometric means ($\widetilde{r_i}$) and fuzzy weights ($\widetilde{w_i}$) of alternatives with respective to "Quality" Criterion

| ALTERNATIVES | $\widetilde{r_i}$ | | | $\widetilde{w_i}$ | | |
|---|---|---|---|---|---|---|
| A1 | 0.265 | 0.281 | 0.303 | 0.052 | 0.063 | 0.078 |
| A2 | 1.000 | 1.186 | 1.442 | 0.198 | 0.265 | 0.371 |
| A3 | 2.621 | 3.000 | 3.302 | 0.519 | 0.672 | 0.850 |
| Total | 3.885 | 4.467 | 5.047 | | | |
| Reverse (power of -1) | 0.257 | 0.224 | 0.198 | | | |
| Increasing Order | 0.198 | 0.224 | 0.257 | | | |

In the last step; the non fuzzy $M_i$ and normalized $N_i$ values are obtained by using centre of area method and shown in Table 10.

Table 10: Averaged and normalized relative weights of each alternative with respect to "Quality" criterion

| ALTERNATIVES | $M_i$ | $N_i$ |
|---|---|---|
| A1 | 0.064 | 0.063 |
| A2 | 0.278 | 0.272 |
| A3 | 0.680 | 0.665 |

Based on these explanations, the normalized non-fuzzy relative weights of each alternative for each criterion are found and tabulated in Table 11.

Table 11: Normalized non-fuzzy relative weights of each alternative for each criterion

| ALTERNATIVES | Quality | Origin | Cost | Delivery | After Sales |
|---|---|---|---|---|---|
| A1 | 0.063 | 0.425 | 0.629 | 0.149 | 0.629 |
| A2 | 0.272 | 0.425 | 0.107 | 0.784 | 0.107 |
| A3 | 0.665 | 0.151 | 0.263 | 0.067 | 0.263 |

By using Table 6 and Table 11, individual scores of each alternative for each criterion are presented in Table 12.

Table 12: Aggregated results for each alternative according to each criterion

| CRITERIA | | Scores of Alternatives with respect to related Criterion | | |
|---|---|---|---|---|
| | Weights | A1 | A2 | A3 |
| Quality | 0.383 | 0.063 | 0.272 | 0.665 |
| Origin | 0.409 | 0.425 | 0.425 | 0.151 |
| **Cost** | 0.074 | 0.629 | 0.107 | 0.263 |
| **Delivery** | 0.057 | 0.149 | 0.784 | 0.067 |
| **After Sales** | 0.076 | 0.629 | 0.107 | 0.263 |
| **Total** | | **0.301** | **0.339** | **0.360** |

Depending on this result, Alternative 3 has the largest total score. Therefore, it is suggested as the best supplier among 3 of them, with respect to 5 criteria and the fuzzy preferences of decision makers.

When this result is compared with the previous study utilizing Fuzzy TOPSIS method to the same case study, the Alternative 3 again outperforms the others (Ayhan, 2013). A1 and A2 result approximately the same values and can be thought as the second best alternatives. However, in this study A2 significantly outperforms A3 and can be thought as the second best supplier.

## 5. CONCLUSION

Depending on various criteria, supplier selection is one of the most important tasks for firms. Since most of these criteria conflict each other, the alternative suppliers should be inspected effectively. Therefore some techniques are developed for this aim. Although there are some more techniques as; TOPSIS, ELECTRE, PROMETHEE, DEMATEL, ANP, etc., in this study Analytical Hierarchy Process technique is used empowered with fuzzy approach. Since the decision makers' preferences depend on both tangible and intangible criteria, these vague linguistic variables should be represented by Fuzzy Set Theory. Hence Fuzzy AHP model is utilized to solve the supplier selection problem of a manufacturing company, which should determine the best supplier among 3 alternatives. These alternative suppliers are inspected with respect to 5 criteria namely; Quality, Origin of the raw material, Cost, Delivery Time, and After Sales Services. As the result of the case study it is seen that the third supplier outperforms the others.

In further studies, as stated before, other models such as Fuzzy ANP or ELECTRE can be applied for the same problem and the results can be compared. In addition, hybrid models combining different methodologies incorporating the strong sides of each can be performed to solve this problem. Furthermore, for more complex problems such as multi sourcing problems, in which no supplier can satisfy all the buyer's requirements, mathematical programming models can be utilized. By using linear programming or goal programming techniques, the decision maker can split order quantities among different suppliers. However since the problem handled in this study, is a single sourcing type, the complicated models are not required to be performed. In conclusion, there are many different types of supplier selection problems to be dealt regarding the supply chain management; several methods can be used for each various type of problem.